\documentclass[letterpaper, 10 pt, conference]{ieeeconf}

\IEEEoverridecommandlockouts                          
\usepackage{amsmath} 
\usepackage{amssymb}  
\usepackage{algorithm}
\usepackage{algpseudocode}
\usepackage{subfig}
\usepackage{graphicx}

\usepackage{multirow}
\usepackage{cuted}
\usepackage[colorlinks = true, urlcolor  = blue]{hyperref}
\usepackage{cite}
\usepackage{booktabs}

\newcommand{\hide}[1]{}

\newcommand{\etal}{\textit{et al.}}
\newcommand\blfootnote[1]{%
  \begingroup
  \renewcommand\thefootnote{}\footnote{#1}%
  \addtocounter{footnote}{-1}%
  \endgroup
}

\title{\LARGE \bf
Robot Synesthesia: In-Hand Manipulation with Visuotactile Sensing
}

\author{
Ying Yuan\textsuperscript{*$2$$\dag$}, 
Haichuan Che\textsuperscript{*$1$},
Yuzhe Qin\textsuperscript{*$1$},
Binghao Huang\textsuperscript{$3$},
Zhao-Heng Yin\textsuperscript{$4$},
\\ Kang-Won Lee\textsuperscript{$5$},
Yi Wu\textsuperscript{$2$},
Soo-Chul Lim\textsuperscript{$5$},
Xiaolong Wang\textsuperscript{$1$}
\thanks{\textsuperscript{$1$} University of California San Diego, CA, USA}
\thanks{\textsuperscript{$2$} Tsinghua University, Beijing, China}
\thanks{\textsuperscript{$3$} University of Illinois Urbana-Champaign, IL, USA}
\thanks{\textsuperscript{$4$} University of California Berkeley, CA, USA}
\thanks{\textsuperscript{$5$} Dongguk University, Seoul, South Korea}
\thanks{\textsuperscript{*} Equal contributions.}
}

\begin{document}
\twocolumn[{%
\renewcommand\twocolumn[1][]{#1}%
\maketitle
\begin{center}
    \vspace{-0.1in}
    \centering
    \captionsetup{type=figure}
    \includegraphics[width=\linewidth]{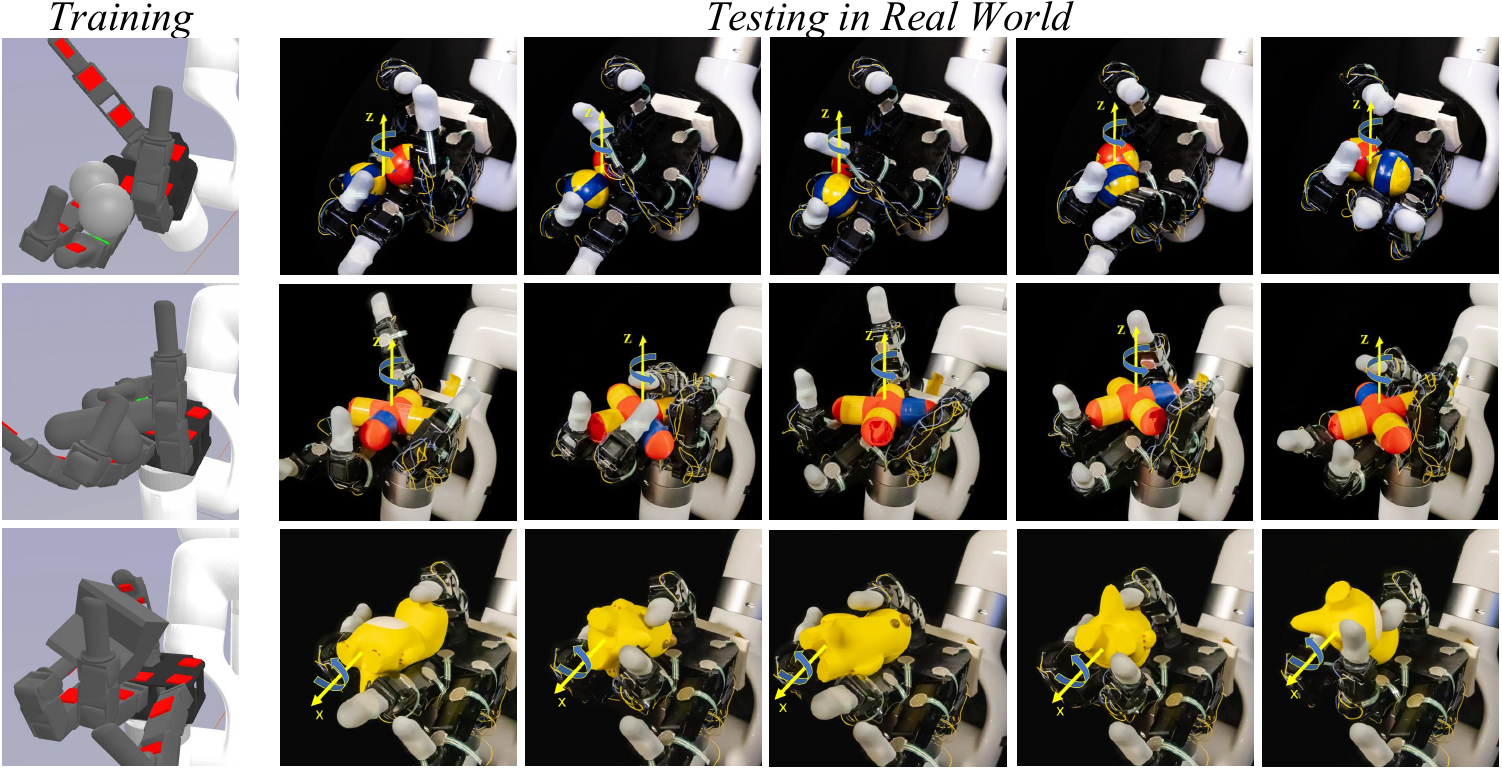}
    \captionof{figure}{We propose \textbf{Robot Synesthesia}, a novel visuotactile approach to perform in-hand object rotation with visual and tactile modalities. We train our policy in simulation on rotating single or multiple objects around a certain axis and then transfer it to the real robot hand without any real-world data.}
    \label{fig:teaser}
    \vspace{-0.1in}
\end{center}
}]

\blfootnote{
\\
\textsuperscript{$1$} University of California San Diego \\
\textsuperscript{$2$} Tsinghua University \\
\textsuperscript{$3$} University of Illinois Urbana-Champaign \\
\textsuperscript{$4$} University of California Berkeley \\
\textsuperscript{$5$} Dongguk University \\
\textsuperscript{*} Equal contributions. \\
\textsuperscript{$\dag$} Work done during internship at UC San Diego.
}

\thispagestyle{empty}
\pagestyle{empty}

\begin{abstract}
Executing contact-rich manipulation tasks necessitates the fusion of tactile and visual feedback. However, the distinct nature of these modalities poses significant challenges. In this paper, we introduce a system that leverages visual and tactile sensory inputs to enable dexterous in-hand manipulation. Specifically, we propose \textbf{Robot Synesthesia}, a novel point cloud-based tactile representation inspired by human tactile-visual synesthesia. This approach allows for the simultaneous and seamless integration of both sensory inputs, offering richer spatial information and facilitating better reasoning about robot actions. 
Comprehensive ablations are performed on how the integration of vision and touch can improve reinforcement learning and Sim2Real performance. Our project page is available at \href{https://yingyuan0414.github.io/visuotactile/}{https://yingyuan0414.github.io/visuotactile/}.
\end{abstract}
\section{Introduction}
\label{sec:intro}
In everyday life, humans effortlessly perform complex manipulation tasks, intuitively using a combination of vision and touch. Considering the intricate task of threading a needle, we begin by visually locating the needle's eye and estimating its size. Holding the needle steady in hand, we use touch information to guide the thread. Our visual sensing guides us in aligning the thread with the needle's eye, while it's the sense of touch from our fingertips that helps us feel the thread's position, even when it's occluded for our eyes to discern accurately. This synergy of vision and touch enables us to interact with our environment with remarkable flexibility and great robustness against occlusion. 

Yet, for robots tasked with similar manipulation tasks, achieving this level of sophistication remains a challenge. There are two primary hurdles that stand in the way of replicating the same level of synergy for robot learning algorithms. (i) \textit{Tactile and visual modality are distinct in nature.} Tactile information is typically sparse and low-dimensional, captured from distinct tactile sensors and provides little contextual details. On the other hand, visual data is dense and high-dimensional, offering a rich tapestry of environmental details. When integrating these two types of data into a single neural network, the model must process and interpret each modality effectively, while also finding a way to synergize this information to facilitate intelligent decision-making. (ii) \textit{Vast amounts of training data are required for such tasks}, which is typically generated within a simulated environment. However, transferring the visuotactile skills learned in a simulator to the real world is a non-trivial problem. Each modality, vision and touch, has its own domain gap. Bridging them concurrently for a combined visual-tactile model even heightens the complexity significantly.

In this paper, we aim to equip the robot with a policy that effectively leverages multi-modal feedback.
In neuroscience studies, certain individuals can perceive color when they touch things, which refers to Tactile-Visual Synesthesia~\cite{simner2012color, davies2013sensational}.
Inspired by it, instead of processing each modality separately in representation learning and merging the learned features later, we propose a novel point cloud-based tactile representation. 
We formulate this representation in a way that ``paints" the tactile data from Force-Sensing Resistor (FSR) in conjunction with the point cloud from the camera into a unified 3D space.
This approach preserves the spatial relationship among the robot links, FSR sensors, and the object being manipulated. Effectively, the robot is equipped to ``see" its tactile interactions, a concept we call \textbf{Robot Synesthesia}.
This method allows for the seamless integration of both sensory inputs from the outset, which offers abundant spatial information, facilitating better reasoning about robot actions. Furthermore, we can easily generate these tactile point clouds in both simulated and real-world using the robot's kinematics. This strategy can reduce the compounding errors of vision and touch during Sim-to-Real transfer, by treating these two modalities as an integrated entity. 

We focus on in-hand object rotation involving one or two objects, and along the x, y, and z axes. The robot is required to interact with a variety of objects via visuotactile sensing, while learning to prevent the objects from slipping off the hand at the same time. 
The task becomes more challenging when rotating two balls (the first row of Figure~\ref{fig:teaser}), due to the high degree of freedom and complex interaction pattern of this double-ball system. Small finger movements are insufficient to rotate them, while excessive motion risks dropping them.
We first train a teacher policy in a physical simulator using Reinforcement Learning (RL) with oracle state information, which is then distilled to a student policy utilizing a PointNet~\cite{pointnet} encoder with visual and tactile inputs. The student policy is then deployed in the real world. 

In our experiments, we evaluate the policy with eight real-world objects. Our policy can solve the challenging double-ball rotation task and generalize to novel objects for the three-axis rotation task. Furthermore, we investigate the critical point sets of the point cloud encoder and show that the proposed tactile representation assists the PointNet in locating critical points, such as fingertips, object surfaces, and tactile points that are crucial for action prediction.

\section{Related Work}
\label{sec:related}
\textbf{Dexterous Manipulation} presents a wealth of opportunities for broad applications~\cite{dex_overview, dex_7, rl-openai, qin2023anyteleop, shaw2023leap, ye2022learning}. 
It enables the execution of intricate manipulative tasks, such as sliding~\cite{sliding_dex, dex_6}, rolling~\cite{rolling_1, dex_3, dex_4, dex_5, rolling_2}, pivoting~\cite{pivot_1,pivot_2}, and regrasping~\cite{regrasp-1, regrasp-2, regrasp-3}.
Earlier methods addressing dexterous manipulation were grounded in classical control~\cite{inhand_1, inhand_4}. However, these methods rely on expert-engineered dynamics models, which restricts their utility for more complex tasks. 
To overcome these limitations, recent research has leveraged deep model-free RL for dexterous manipulation~\cite{corl21-inhand, qin2022dexpoint, khandate2022feasibility, bao2023dexart, huang2023dynamic}.
Further enriching these advancements, imitation learning and combining common RL with demonstrations has led to higher sample efficiency and more natural manipulation behaviors~\cite{qin2022dexmv, dapg, arunachalam2022dexterous, qin2022one, wu2023learning, liu2022herd, patel2022learning, arunachalam2022holo, sivakumar2022robotic}. 
Dexterous in-hand manipulation has been a focal point of research interest recently~\cite{inhand_2,inhand_3,inhand_5, qi2023hand, yin2023rotating, extreme, rostel2023estimator}. 
To generalize to new objects, researchers have explored different sensors to capture object geometry and dynamic properties.
Qi~\etal\cite{qi2023hand} demonstrated that policy could infer object position and physical properties using proprioceptive history. But without explicit object information, it was only effective for z-axis rotation tasks. Yin~\etal~\cite{yin2023rotating} proposed integrating binary tactile signals with proprioception for this task. However, the tactile signal was too sparse to capture detailed geometric attributes and thus could not handle objects with non-convex shapes. To solve this, Chen~\etal~\cite{chen2023visual} utilized depth information to aid object rotation while Guzey~\etal~\cite{guzey2023dexterity} learned non-parametric policies on both vision and touch signal.
Most similar to us, a recent study~\cite{qi2023general} utilizes RGB images from optical tactile sensors and depth images for in-hand rotation. However, it necessitated continuous contact between the object and tactile sensors, constraining it to smaller objects that can be rotated on the fingertips. In contrast, our work does not impose any specific requirements on the object's initial location and can handle objects of diverse shapes and sizes. Furthermore, our tactile point cloud representation provides explicit 3D information about the object and tactile sensors' location, while their models are based on 2D images. As a result, our policy can solve tasks requiring more complex 3D spatial reasoning, such as rotating two balls simultaneously.

\textbf{Visuotactile Manipulation},
the integration of visual and tactile modalities, is a fundamental mechanism for human interaction with the environment~\cite{jenmalm1997visual}, which presents significant potential for enhancing robot manipulation capabilities~\cite{chen2022visuo, falco2017cross, wang20183d, billard2019trends, guzey2023dexterity}. The visual modality offers a comprehensive, non-contact perspective of the environment, while the tactile modality complements this by offering detailed, contact-dependent properties such as texture, temperature, hardness, and weight. 
The key to integrating these modalities in robotic manipulation lies in two aspects: (i) the representation of each modality, and (ii) the strategy employed to fuse these modalities.
In the visual modality, RGB images are a common choice due to their widespread availability~\cite{falco2017cross, calandra2018more}. But these images do not inherently capture distance information, which is often critical in manipulation tasks. To address this limitation, researchers~\cite{qi2023general, chen2023visual} proposed using depth to handle more contact-intensive tasks such as in-hand rotation. Different from these methods, our approach utilizes the point cloud captured by a camera, inherently incorporating 3D information into the visual representation.
For the tactile modality, raw sensor readings are a natural choice~\cite{lee2019making, bischoff1999integrating, guzey2023see}.
However, for a smaller Sim-to-Real gap, the binary contact vector has been used~\cite{yin2023rotating, bin_contact_1, bin_contact_2, liang2020hand}. Notably, vision-based tactile sensors, such as DIGIT~\cite{lambeta2020digit} and GelSighT~\cite{yuan2017gelsight}, can also encode tactile information into RGB images.
Another critical consideration is the design of a multi-modal learning paradigm. Most existing approaches favor combining modalities at the feature level, where separate feature extractors are trained for each modality, and the predicted features are concatenated as a multi-modal representation~\cite{lee2019making, hansen2022visuotactile}. When utilizing optical tactile sensors with tactile images, the combination can also be performed at the input level, as both vision and touch are represented as RGB images~\cite{wang20183d, calandra2018more}. In contrast, our method represents tactile data as a point cloud and merges with the camera point cloud at the input level. This approach treats the combined visual and tactile data as a single input to the policy network. This innovative design, which we term Robot Synesthesia, enriches the contextual understanding for both modalities and encodes the sensory data into a cohesive 3D space.

\section{Visuotactile Dexterous Manipulation}
\label{sec:system}
\subsection{System Setup}
\begin{figure}[]
    \centering
    \includegraphics[width=0.8\linewidth]{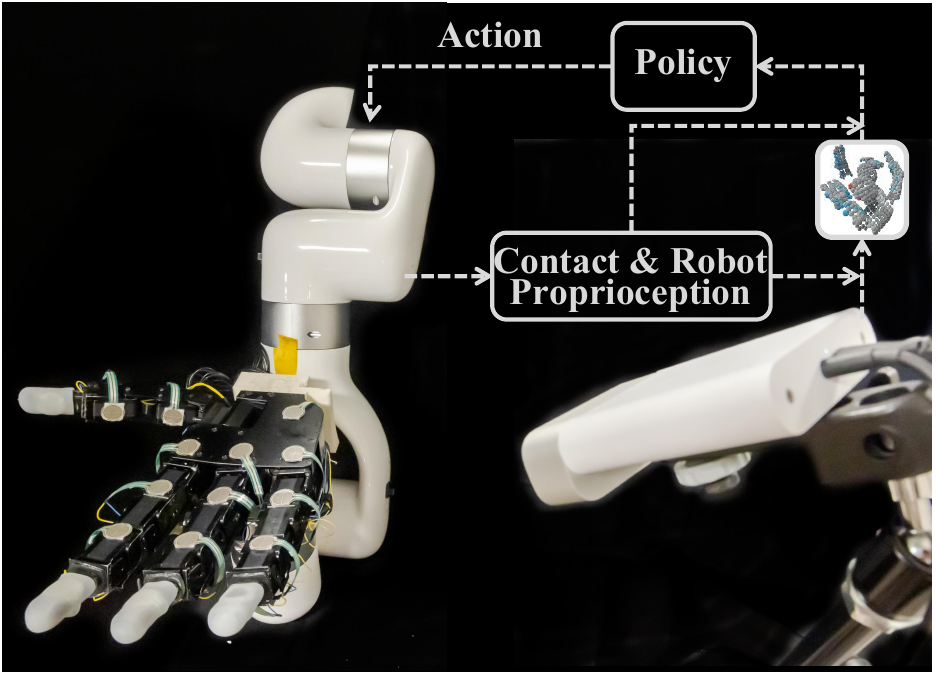}
    \caption{\textbf{Real-World Setup.} We use an Allegro Hand attached with 16 Force-Sensing Resistors. A Microsoft Azure Kinect camera is placed facing forward the robot.}
    \label{fig:real-setup}
    \vspace{-6mm}
\end{figure}

As is shown in Figure~\ref{fig:real-setup}, our hardware setup consists of an XArm6 robot arm and a 16-DOF Allegro Hand with a depth camera. We attach 16 Force-Sensing Resistors (FSR) as tactile sensors to the palm and finger links of the robot hand as suggested by \cite{yin2023rotating}. We gather the contact signal from each sensor, then binarize the measurement according to a predetermined threshold $\theta_{th}$ to abridge the Sim-to-Real gap. We use Isaac Gym\cite{isaacgym} as the rigid body physics simulator. The simulation setup is visualized in Figure~\ref{fig:teaser}. 

\begin{figure}
    \centering
    \includegraphics[width=\linewidth]{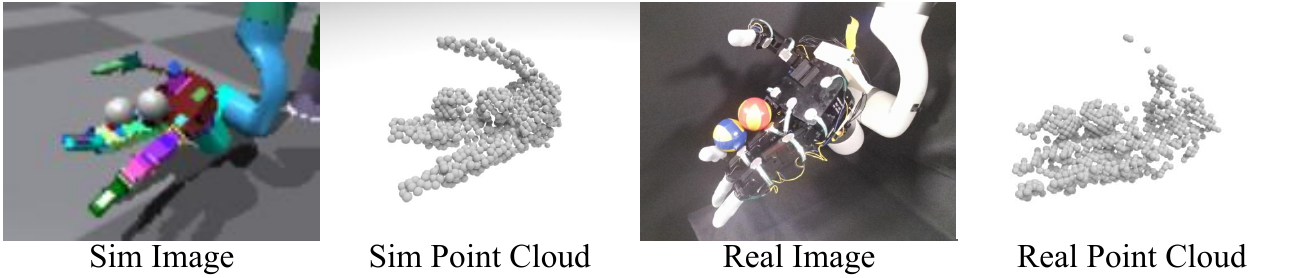}
    \caption{\textbf{Point Cloud Visualization in Sim and Real.} The Sim-to-Real gap is notably larger for RGB images compared to point clouds, leading us to select point clouds as the visual observation for our policy.
    }
    \label{fig:sim2real-gap}
    \vspace{-5mm}
\end{figure}

To obtain visual observations, we place a Microsoft Azure Kinect camera beside the hand in both real and simulated settings. 
We then generate the point cloud using the depth image. 
As illustrated in Figure~\ref{fig:sim2real-gap}, the point clouds in simulation closely mirror those in reality, especially compared with RGB images. We create an augmented point cloud~\cite{qin2022dexpoint} from the robot's proprioception to model the spatial relationship between the hand and the object, by sampling on the robot's mesh at the current pose. To differentiate between the camera-generated point cloud and the augmented one, we append a one-hot vector to each point. Point cloud visualizations are shown in Figure~\ref{fig:method:teacher-student-pipeline}.
The control frequency remains consistent at 10Hz in both simulated and real environments.

\subsection{Benchmark Problems}
We study the dexterity of Robot Synesthesia through an in-hand object rotation task, where the goal is manipulating one or more hand-held objects to rotate along a specific axis. In this paper, we mainly focus on three distinct benchmark problems:
\textbf{(i) Wheel-Wrench Rotation}: Inspired by scenarios where a user must switch handles on a wrench during use, this task involves rotating an artificial multi-way wheel wrench along the z-axis in hand without dropping it. To successfully complete this task, the robot must visually identify the next "possible" handle for interaction while concurrently sensing the wrench's rotation via touch.
\textbf{(ii) Double-Ball Rotation}: This task requires the simultaneous manipulation of two identical balls to rotate around each other along the z-axis. Given that tactile feedback alone cannot distinguish between the balls, it is crucial for the robot to visually locate both.
\textbf{(iii) Three-Axis Rotation}: This task extends beyond the z-axis to require the robot to rotate objects around a fixed x or y-axis. Moreover, the policy should demonstrate the ability to manipulate a variety of objects with different shapes, extending its dexterity to objects not included in the training set.

\section{Learning Visuotactile Dexterity}
\label{sec:method}
\begin{figure*}
    \centering
    \includegraphics[width=0.88\linewidth]{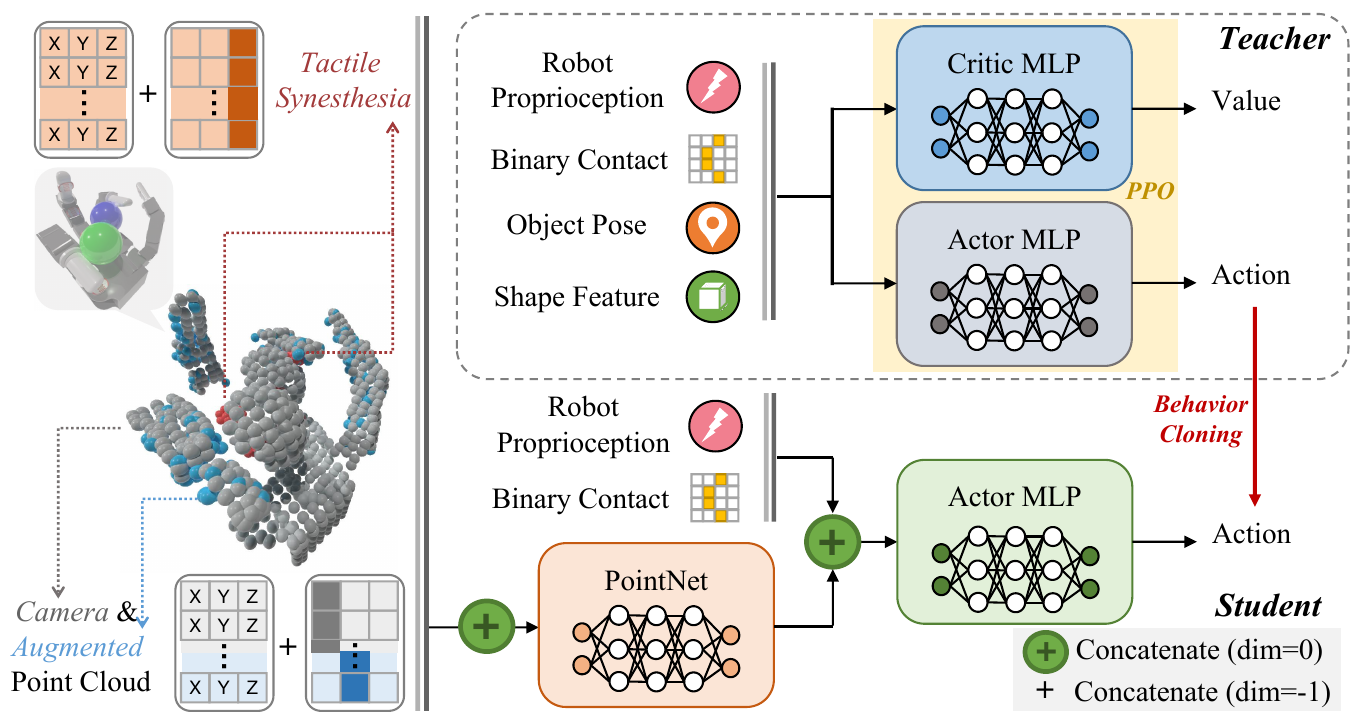}
    \caption{\textbf{Training Pipeline.} Our teacher policy takes robot proprioception, binary contact, object pose, and object shape embedding as input. After training the teacher policy via RL, we distill it to a visuotactile-based student policy. Besides robot proprioception and touch signal, the student policy takes a point cloud from depth-camera, an augmented point cloud based on robot proprioception, and the proposed tactile point cloud. We use one-hot vectors to distinguish point clouds. Note that we've eliminated noise from the point clouds for better clarity here.
    }
    \label{fig:method:teacher-student-pipeline}
    \vspace{-5mm}
\end{figure*}

\subsection{Problem Formulation}
We formulate the in-hand object rotation task as a Markov Decision Process $(\mathcal{S}, \mathcal{A}, \mathcal{P}, \mathcal{R}, \gamma)$. Here, $\mathcal{S}$ is the state space, $\mathcal{A}$ is the action space, $\mathcal{P}(s'|s, a)$ is the transition probability, $\mathcal{R}$ is the reward function, and $\gamma$ is the discounted factor. \hide{The robot agent observes $s_t$ at time step $t$ and takes action $a_t=\pi_{\theta}(s_t)$ calculated by the current policy $\pi_{\theta}$, and then receives reward $r_t=\mathcal{R}(s_t, a_t, s_{t+1})$.} The objective is to find an optimal $\theta^*$ that maximizes the expected accumulated reward $\sum_{t=0}^T \gamma^t r_t$. 
An episode terminates when reset conditions are achieved or the agent reaches the maximum number of steps $T$. We prune unnecessary explorations when the object falls off the hand \hide{or tips over along an axis other than the rotation axis }to facilitate efficient training. 
\subsubsection{State}
The state of the system consists of the joint position $q_t\in \mathbb{R}^{16}$ of the Allegro hand, the binary tactile signal $o_t\in \{0, 1\}^{16}$, the rotation axis $k\in\mathbb{S}^2$, the previous position target $\hat{q}_t\in\mathbb{R}^{16}$, the camera point cloud $P^{c}_{t}\in \mathbb{R}^{N_{c}\times 3}$, the augmented point cloud $P^{a}_{t}\in\mathbb{R}^{N_{a}\times 3}$, and the tactile point cloud $P_t^{touch}\in\mathbb{R}^{N_{a}\times 3}$.
\subsubsection{Action}
At each step, the action provided by the policy network is a relative control command $a_t\in\mathbb{R}^{16}$ and a PD controller drives the robot hand to approach $\hat{q}_{t+1}=\hat{q}_t+\hat{a}_t$. Note that we employ an exponential moving average in our implementation, i.e., $\hat{a}_t=\eta a_t+(1-\eta)\hat{a}_{t-1}, t\geq 1$ and let $\hat{a}_0=0$. We set $\eta=0.8$ in our experiments.
\subsubsection{Reward}
We design a reward function for robust and transferable in-hand rotation, which is a weighted composition of several components: 
\begin{equation}
r_t=c_1r_{rot}+c_2r_{vel}+c_3r_{dist}+c_4r_{torq}+c_5r_{work}+c_6r_{ctrl}.
\label{rew-equation}
\end{equation}
$r_{rot}$ rewards the object's rotation angle. $r_{vel}$ penalizes the object's linear velocity to discourage motions that translate the object. $r_{dist}$ is a decreasing function regarding the distance between the object and the fingertips, encouraging the fingers to approach the object in hand and interact with it. $r_{torq}$ penalizes large torques, $r_{work}$ penalizes the work of the controller, and $r_{ctrl}$ penalizes the control error between command targets and real robot motion. We additionally implement a large penalty when the object falls off the hand. $c_1, c_2,\cdots, c_6$ are tuned hyper-parameters.

\subsection{Tactile-Visual Synesthesia}
Instead of processing tactile and visual modalities separately for feature extraction, we unify tactile and visual modalities by projecting them onto a single 3D space, similar to how humans might simultaneously perceive touch and visual stimuli in their minds. 
Concretely, for each tactile sensor on the hand that detects a signal (i.e., $o_{t, i}=1$), we sample points on the sensor's meshes to create a tactile-based point cloud $P_t^{touch}$. When combining $P_t^{touch}$ with $P_t^c$ and $P_t^a$, we provide the policy network with a spatial relationship of all the observation entities. In our implementation, we set the number of sampled points $N_c=512, N_a=8n_{link}$, and $N_t=8n_{touch}$, where $n_{link}=21$ is the number of links on hand and $n_{touch}\in \{0, \cdots, 16\}$ is the number of triggered tactile sensors. Our experiments demonstrate that points sampled from active tactile sensors are implicitly chosen by our learned policy for representation learning. Note that we transform all point clouds to the hand palm~\cite{liu2022frame} frame before feeding them into the neural network.

\subsection{Teacher-Student Training Pipeline}
Learning the controller $\pi$ using RL is data inefficient when the observation is high-dimensional, e.g., point clouds\hide{, because the policy is required to extract essential information from high-dimensional observations and discern which actions are high-rewarding}. To mitigate this issue, we employ a teacher-student learning approach to obviate training vision policies with RL, as shown in Figure~\ref{fig:method:teacher-student-pipeline}.
\subsubsection{Teacher policy training}
We first use the proximal policy optimization (PPO)\cite{ppo} to train teacher policies with low-dimensional states. Its input consists of the joint position of the Allegro hand $q_t$, the binary tactile signal $o_t$, the rotation axis $k$, the previous position target $\hat{q}_t$, the object's position $x_t\in\mathbb{R}^3$, its velocity $v_t\in\mathbb{R}^3$, its angular velocity $w_t\in\mathbb{R}^3$, and the object's shape feature embedding $f\in\mathbb{R}^{32}$. For tasks that require generalizability over multiple objects, we encode the shape information via a pre-trained PointNet\cite{pointnet} encoder in \cite{wu2023learning}. Note that for each object, the shape feature embedding remains the same throughout the training.
Given the state information, we use a Multi Layer Perceptron (MLP) for both policy and value networks. We stack the current state with 3 historical states as input for better perception.

\subsubsection{Student policy training}
After using RL to train the teacher policy, we distill it to a student policy with visuotactile input. Concretely, its input includes the joint position $q_t$, the binary tactile signal $o_t$, the rotation axis $k$, and the previous position target $\hat{q}_t$. We stack it with 3 historical states. For the visual observation, the input includes camera point cloud $P_t^c$, augmented point cloud $P_t^a$, and the proposed tactile point cloud $P_t^{touch}$. We attach a one-hot vector to each point and concatenate them together as $P_t$.

We use PointNet\cite{pointnet} as the point cloud encoder and feed the latent vector and other inputs into an MLP. We adopt a two-stage distillation pipeline: We first collect a teacher dataset $\mathcal{D}$ of $5120k$ transitions and use Behavior Cloning (BC) to pre-train our student policy network; in the second stage, we use Dataset Aggregation (DAgger)\cite{dagger} to fine-tune the network for more robust behavior.

\section{Experiments}
\label{sec:exp}
In this part, we compare our robot synesthesia approach to several baselines in both the simulation and the real. Specifically, we are interested in the following questions:
\begin{itemize}
    \item [1)]
    \textit{How much benefit do visual and tactile modalities offer?}
    \item [2)]
    \textit{Is teacher-student pipeline necessary for efficient training given both tactile and visual modalities?}
    \item [3)]
    \textit{How does our policy network process the two modalities through synesthesia?}
\end{itemize}

\subsection{Setup}
\subsubsection{Object Dataset}
\begin{figure}[]
    \centering
    \includegraphics[width=\linewidth]{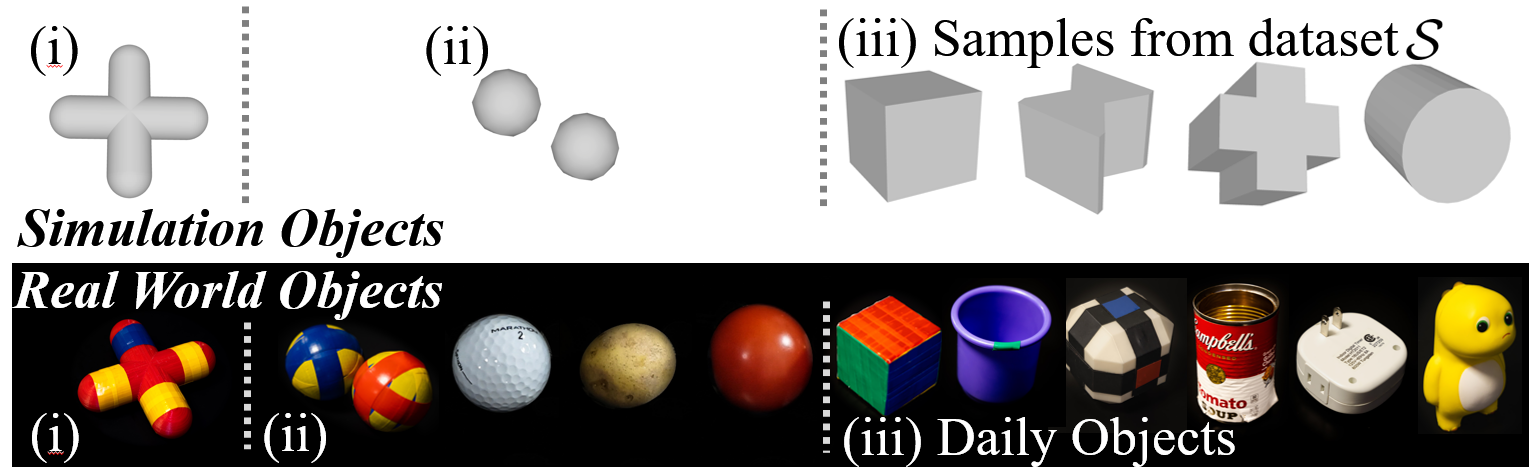}
    \caption{\textbf{Object Sets in Sim and Real.} We use artificial objects for training and daily objects for testing.}
    \label{fig:obj}
    \vspace{-3mm}
\end{figure}

The objects used for the benchmark problems are shown in Figure~\ref{fig:obj}. For task (i), we use an artificially designed four-way wheel wrench in both simulation and real. For task (ii), we use two balls of the same size. For task (iii), we train and evaluate our policy on a set of artificial objects of common geometries $\mathcal{S}$, such as cuboids, cylinders, polygons, etc. in the simulation. For real deployment, we use cubes and other daily objects of distinct sizes and shapes to test our policy's generalization ability.

\subsubsection{Evaluation Metric}
To evaluate the policy performance, we use the following metrics as suggested by~\cite{qi2023hand}. 
\begin{itemize}
    \item [1)]
    \textbf{Cumulative Rotation Reward (CRR)} is the reward our agent obtains in an episode. We use it to evaluate the rotation capability of a policy in the simulation. 
    \item [2)]
    \textbf{Cumulative Rotation Angle (CRA)} is the angle (by rounds) the object rotates along the axis in an episode. We use it to evaluate a policy in the real.
    \item [3)]
    \textbf{Time-to-Fall (TTF/Duration)} is the length of an episode (by seconds). TTF varies when the object falls off before the maximal episode length.
\end{itemize}

\begin{figure}[]
    \centering
    \includegraphics[width=\linewidth]{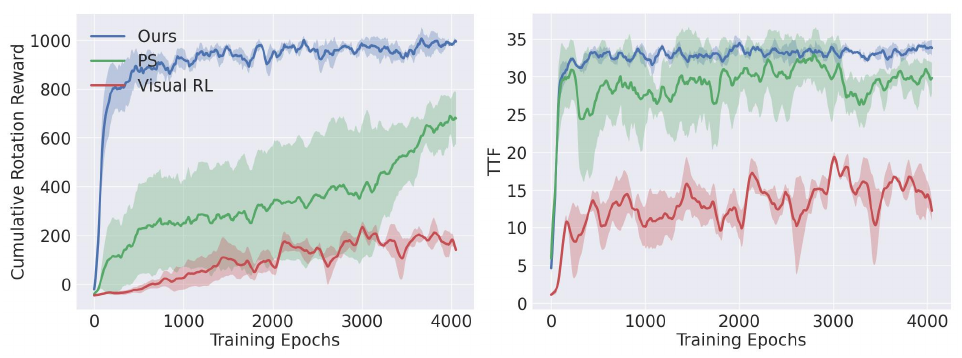}
    \caption{Learning curve of teacher policy on double-ball rotation. The results are averaged on 3 seeds.
    }
    \label{fig:exp:curve}
    \vspace{-3mm}
\end{figure}
\begin{figure}[]
    \centering
    \includegraphics[width=\linewidth]{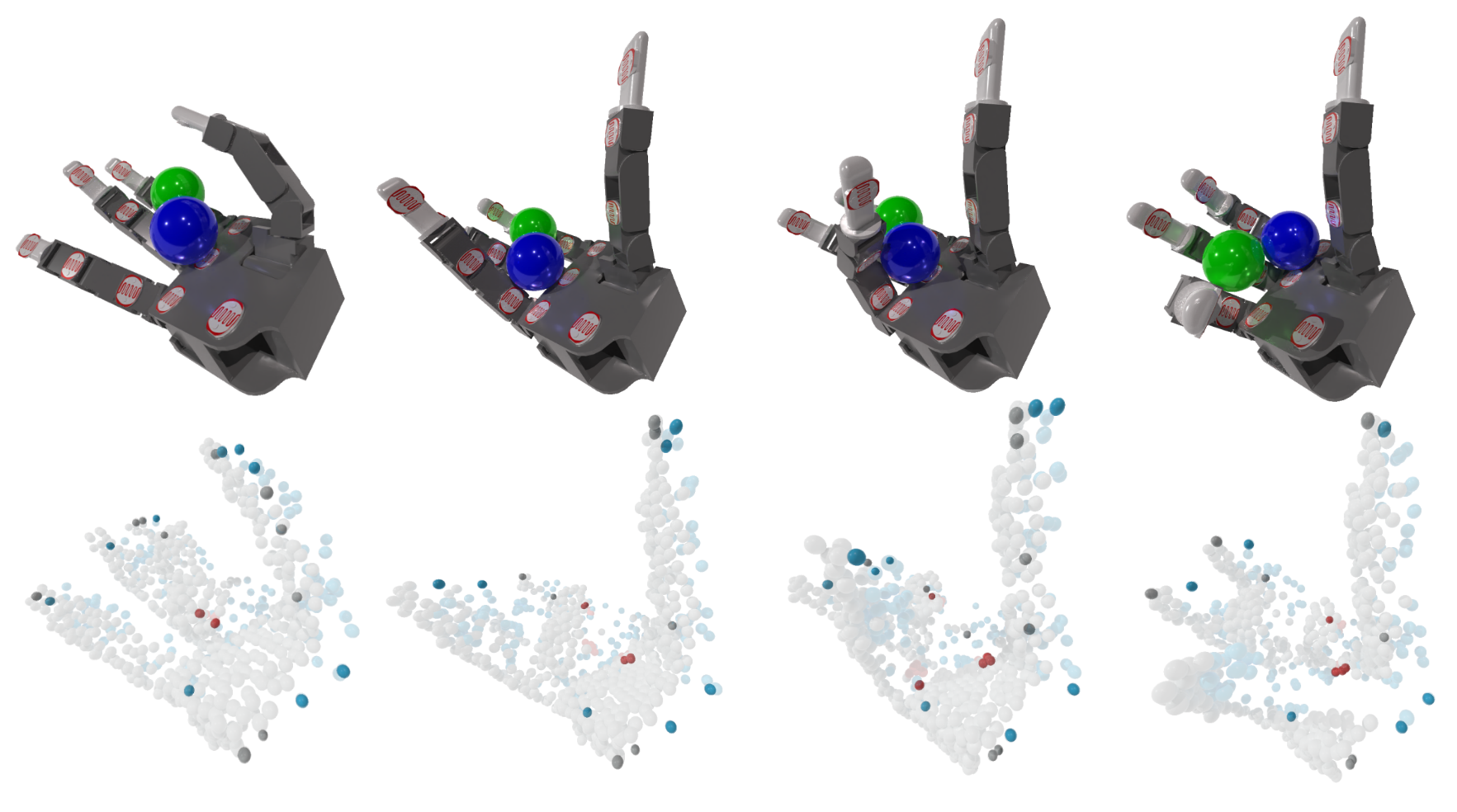}
    \caption{Visualization of selected point clouds (foreground) among observed point clouds (background) during evaluation. Red points belong to the proposed tactile point cloud.}
    \label{fig:pointnet}
    \vspace{-5mm}
\end{figure}

\subsection{Stage I: RL training with different sensing capabilities}

\begin{table*}[htbp]
\centering
\caption{Evaluation of RL policies of different sensing capabilities on three benchmark problems in the simulation. Each policy is tested for 500 episodes. The results are averaged over 3 policies trained on 3 seeds. Each trial lasts 50 seconds.}
\resizebox{1.02\linewidth}{!}{
\begin{tabular}{ccccccccccccccc}
\toprule[1pt]
&  \multicolumn{2}{c}{4-way Wrench} \hide{& \multicolumn{2}{c}{5-way Wrench}} &  \multicolumn{2}{c}{Double Balls}& \multicolumn{2}{c}{Multi-Object (x-axis)}& \multicolumn{2}{c}{Multi-Object (y-axis)}& \multicolumn{2}{c}{Multi-Object (z-axis)}\\
\multirow{-2}{*}{Obs Type} & CRR& TTF\hide{& CRR & TTF} &CRR & TTF& CRR& TTF & CRR & TTF& CRR & TTF\\ \midrule[0.5pt]
Visual RL  & 10.9$_{\pm 2.2}$  & 8.1$_{\pm3.2}$ \hide{& 18.4$_{\pm 13.8}$  & 7.0$_{\pm 0.9}$}& 127.8$_{\pm 78.6}$ & 10.5$_{\pm 3.7}$ & 15.3$_{\pm 8.2}$ & 16.8$_{\pm 11.8}$ & 22.4$_{\pm 8.8}$ & 21.4$_{\pm 17.8}$& 29.5$_{\pm 7.1}$ & 2.9$_{\pm 0.4}$\\
PS & 440.7$_{\pm 590.3}$ & 22.6$_{\pm18.5}$ \hide{& 665.8$_{\pm 483.1}$& 35.4$_{\pm 16.7}$}& 620.9$_{\pm 39.9}$ & 28.8$_{\pm 0.7}$ & 446.1$_{\pm 137.7}$ & 33.1$_{\pm 7.1}$ & 552.1$_{\pm 318.7}$ & 33.5$_{\pm 8.3}$ & 878.7$_{\pm 528.3}$ & 36.9$_{\pm 15.4}$\\
Ours & \textbf{1011.1}$_{\pm329.9}$ & \textbf{47.5$_{\pm0.4}$} \hide{& \textbf{892.8}$_{\pm88.4}$ & \textbf{44.1}$_{\pm 1.9}$} & \textbf{1045.3}$_{\pm 64.9}$ & \textbf{36.2}$_{\pm 2.3}$ & \textbf{985.9}$_{\pm 174.1}$ & \textbf{45.1}$_{\pm 2.6}$ & \textbf{987.3}$_{\pm 141.9}$ & \textbf{46.8}$_{\pm 1.0}$ & \textbf{1353.7}$_{\pm 123.8}$ & \textbf{48.2}$_{\pm 0.4}$ \hide{\\ \midrule[0.5pt]
& \multicolumn{2}{c}{Multi Object (x-axis)}& \multicolumn{2}{c}{Multi Object (y-axis)}& \multicolumn{2}{c}{Multi Object (z-axis)}\\
\multirow{-2}{*}{Obs Type} & CRR& TTF & CRR & TTF& CRR & TTF \\ \midrule[0.5pt]
Visual RL & 15.3$_{\pm 8.2}$ & 16.8$_{\pm 11.8}$ & 22.4$_{\pm 8.8}$ & 21.4$_{\pm 17.8}$& 29.5$_{\pm 7.1}$ & 2.9$_{\pm 0.4}$\\
PS & 446.1$_{\pm 137.7}$ & 33.1$_{\pm 7.1}$ & 552.1$_{\pm 318.7}$ & 33.5$_{\pm 8.3}$ & 878.7$_{\pm 528.3}$ & 36.9$_{\pm 15.4}$\\
Ours & \textbf{985.9}$_{\pm 174.1}$ & \textbf{45.1}$_{\pm 2.6}$ & \textbf{987.3}$_{\pm 141.9}$ & \textbf{46.8}$_{\pm 1.0}$ & \textbf{1353.7}$_{\pm 123.8}$ & \textbf{48.2}$_{\pm 0.4}$}\\ \bottomrule[1pt]
\end{tabular}
}
\label{Table:rl-eval}
\end{table*}

\begin{table*}[htbp]
\centering
\caption{Evaluation of student policies of different sensing capabilities on three benchmark problems in the simulation. Each policy is tested for 500 episodes. Each trial lasts 50 seconds.}
\begin{tabular}{ccccccccccccccccccc}
\toprule[1pt]
&  \multicolumn{2}{c}{4-way Wrench} \hide{& \multicolumn{2}{c}{5-way Wrench}} &  \multicolumn{2}{c}{Double Balls} & \multicolumn{2}{c}{Multi-Object (x-axis)}& \multicolumn{2}{c}{Multi-Object (y-axis)}& \multicolumn{2}{c}{Multi-Object (z-axis)}\\
\multirow{-2}{*}{Obs Type} & CRR& TTF\hide{& CRR & TTF} &CRR & TTF& CRR& TTF & CRR & TTF& CRR & TTF\\ \midrule[0.5pt]
Touch & 363.2& 23.6 \hide{& 18.8 & 10.4} & 317.1 & 13.6 & 390.9 & 24.2 & 710.9 & 42.6 & 702.4 & 35.6\\
Cam+Aug  & 94.6 & 15.2 \hide{& 27.5 & 5.6} & 162.7 & 9.6 & 630.9 & 40.3 & \textbf{743.5} & \textbf{42.9} & 624.2 & 29.2\\
Touch+Cam+Aug & 344.1 & 21.1 \hide{& 185.9 & 20.8} & 148.6 & 9.6 & \textbf{881.1} & \textbf{47.4} & 619.0 & 41.3 & 909.8 & 37.7 \\
Touch+Cam+Aug+Syn & \textbf{504.0} & \textbf{29.2} \hide{& \textbf{202.9} & \textbf{21.1}} & \textbf{407.7} & \textbf{17.1} & 846.9 & 39.9 & 686.8 & 41.2 & \textbf{1035.0} & \textbf{41.3} \hide{ \\  \midrule[0.5pt]
& \multicolumn{2}{c}{Multi Object (x)}& \multicolumn{2}{c}{Multi Object (y)}& \multicolumn{2}{c}{Multi Object (z)}\\
\multirow{-2}{*}{Obs Type} & CRR& TTF & CRR & TTF& CRR & TTF \\ \midrule[0.5pt]
$\times$Tactile & 630.9 & 40.3 & \textbf{743.5} & \textbf{42.9} & 624.2 & 29.2\\
$\times$Pointcloud& 390.9 & 24.2 & 710.9 & 42.6 & 702.4 & 35.6\\
$\times$Synesthesia & \textbf{881.1} & \textbf{47.4} & 619.0 & 41.3 & 909.8 & 37.7\\
Ours & 846.9 & 39.9 & 686.8 & 41.2 & \textbf{1035.0} & \textbf{41.3}}\\  \bottomrule[1pt]
\end{tabular}
\label{Table:student-eval}
\end{table*}

\begin{table*}[htbp]
\centering
\caption{Evaluation of policies (CRA/TTF) in the real-world deployment. The above two lines refer to non-visual methods and the rest are visual policies. Each policy is tested for 5 episodes. Each trial lasts 60 seconds.}
\begin{tabular}{ccccccccccccccccccc}
\toprule[1pt]
Obs Type &  \\
(CRA/TTF)& \multirow{-2}{*}{4-way Wrench} &   \multirow{-2}{*}{Double Balls}& \multirow{-2}{*}{Multi-Object (x-axis)}& \multirow{-2}{*}{Multi-Object (y-axis)}& \multirow{-2}{*}{Multi-Object (z-axis)} \\
\midrule[0.5pt]
Non-visual RL  &  0.25/60.0 & 0.2/28.6 & 0.35/60.0 & 1.0/60.0 & 8.6/60.0\\
Touch & 0.25/60.0 & 15.6/26.7 & 0.7/60.0 & 0.2/60.0 & 7.4/60.0\\  \midrule[0.5pt]
Cam+Aug &  0.25/60.0 & 10.1/20.8 & 0.25/60.0 & 1.0/33.3 & 5.1/60.0\\
Touch+Cam+Aug & 0.25/60.0 & 18.8/32.7 & 0.5/60.0 & \textbf{1.4/28.3} & 5.1/57.1 \\
Touch+Cam+Aug+Syn & \textbf{1.5/43.0} & \textbf{22.9/36.6}& \textbf{2.1/26.6} & 0.9/29.3 & \textbf{10.2/60.0} \\ \bottomrule[1pt]
\end{tabular}
\label{Table:real-eval}
\vspace{-5mm}
\end{table*}

In this section, we experiment with training RL teacher policies in the simulation. We compare our implementation with two baselines: \textbf{Partially-observable-State(PS/Non-visual RL)} policy\cite{yin2023rotating} is a non-visual policy that observes only robot proprioception and contact signals; \textbf{Visual RL} policy is a visuotactile policy trained via RL from scratch. Figure \ref{fig:exp:curve} shows learning curves of double-ball rotation and multi-object rotation around the x-axis. We find that our method achieves a higher reward compared with PS, and that visual RL hardly learns high-rewarding actions within the same number of training epochs. We evaluate policies trained on the benchmark problems for 500 episodes as is shown in Table \ref{Table:rl-eval}. Our approach outperforms PS and Visual RL in all the tasks. This indicates that the ground-truth object pose is essential for robust and meticulous manipulation, especially when the object, e.g. multi-way wrenches, requires different actions as its direction varies. Also, learning vision-based RL policies is data-inefficient, probably because the policy needs to extract features from high-dimensional inputs and learn high-rewarding actions simultaneously. 

\subsection{Stage II: Imitation learning with different sensing capabilities}
In this stage, we distill teacher policies to visuotactile policies and perform ablation study for different sensing capabilities. As shown in Table~\ref{Table:student-eval}, \textbf{Touch} refers to binary contact, \textbf{Cam} refers to camera-based point clouds, \textbf{Aug} refers to augmented point clouds, and \textbf{Syn} refers to proposed tactile point clouds. For rotating objects of regular shapes, visual policies achieve similar dexterity to each other. However, when it comes to more challenging objects like multi-way wrenches and two balls, our method outperforms all the baselines. 

\subsection{Real-world Deployment}
We transfer the visuotactile policies to the real robot without any fine-tuning and test whether visual policies continue to provide benefits. The results are shown in Table~\ref{Table:real-eval}. Although visual policy might show comparable performance for simple geometry in simulation, the advantage of integrating vision and touch becomes more significant when deployed to the real world. \textbf{These results highlight the low domain gap of our proposed tactile point cloud representation.}
A rudimentary concatenation of tactile signals and the extracted features of point clouds could increase the challenge for the policy to comprehend their underlying relationship. In contrast, our proposed visual-tactile synesthesia approach generally offers benefits. We also observe that visual policies tend to operate more cautiously, making occasional adjustments to nudge the object back to the palm center, while policies lacking visual perception tend to execute an almost fixed sequence of motion, irrespective of the object's deviation or instances of it becoming stuck.

\subsection{Qualitative Analysis: Visualization of PointNet intermediates}
In PointNet, the input point cloud is fed into a local MLP extracting features of each point before a Max Pooling layer over points in each dimension of the features.
Thus, PointNet is trained to implicitly select no more than $c_{out}$ points for representation learning, where $c_{out}$ is the output dimension of PointNet.
We visualize in Figure~\ref{fig:pointnet} the points selected by our policy during evaluation. Interestingly, we find that our policy uses 42.7\% of tactile-based points on average and the rest points are mainly from the tips or edges of fingers and the palm. This indicates that the point cloud encoder can extract meaningful features based on our visual-tactile synesthesia design.

\section{Conclusion}
\label{sec:conclusion}
This paper introduces a system for in-hand dexterous manipulation utilizing visuotactile sensing. We propose a novel tactile representation based on point clouds, and a paired network architecture to leverage it. Our results show that the policy, which has been trained in a simulator using vision and touch input, effectively transfers to the real world. It can solve complex tasks such as double-ball rotation and generalize to novel objects. Future work may encompass goal-conditioned object rotation tasks and the integration of optical tactile sensors. We are committed to releasing the code for our simulation environment and training pipeline.

\section{Acknowledgement}
Acknowledgment: This work was supported, in part, by the Qualcomm Innovation Fellowship, and the Technology Innovation Program (20018112, Development of autonomous manipulation and gripping technology using imitation learning based on visual and tactile sensing) funded by the Ministry of Trade, Industry \& Energy (MOTIE), Korea.

\bibliographystyle{IEEEtran}
\bibliography{IEEEabrv, main}



\end{document}